\documentclass{Interspeech}

\usepackage{cite}
\usepackage{amsmath,amssymb,amsfonts}
\usepackage{algorithmic}
\usepackage{graphicx}
\usepackage{textcomp}
\usepackage{xcolor}
\usepackage{acronym}
\usepackage{booktabs}
\usepackage{multirow}
\usepackage{hyperref}
\usepackage{cleveref}
\usepackage{xcolor}
\usepackage{colortbl}
\usepackage{amssymb}

\usepackage{tablefootnote}
\usepackage{setspace}

\interspeechcameraready

\acrodef{nar}[NAR]{Non-Autoregressive}
\acrodef{ar}[AR]{Autoregressive}
\acrodef{dac}[DAC]{Diffusion-based Audio Captioning}
\acrodef{ddpm}[DDPM]{Denoising Diffusion Probabilistic Model}
\acrodef{mse}[MSE]{Mean Squared Error}
\acrodef{ce}[CE]{Cross Entropy}
\acrodef{nlp}[NLP]{Natural Language Processing}
\acrodef{tps}[TPS]{Tokens Per Second}
\acrodef{aps}[APS]{Audios Per Second}
\acrodef{llm}[LLM]{Large Language Model}

\definecolor{LightCyan1}{rgb}{0.88,1,1}

\def\BibTeX{{\rm B\kern-.05em{\sc i\kern-.025em b}\kern-.08em
    T\kern-.1667em\lower.7ex\hbox{E}\kern-.125emX}}

\title{Towards Diverse and Efficient Audio Captioning via Diffusion Models}

\author[affiliation={1,2}]{Manjie}{Xu$^{*,}$}
\author[affiliation={1}]{Chenxing}{Li$^{*, \dagger,}$}
\author[affiliation={3}]{Yong}{Ren}
\author[affiliation={4}]{Xinyi}{Tu}
\author[affiliation={3}]{Ruibo}{Fu}
\author[affiliation={2}]{Wei}{Liang$^{\dagger,}$}
\author[affiliation={5}]{Dong}{Yu$^{\dagger,}$}

\affiliation{}{Tencent AI Lab, Beijing}{China}
\affiliation{}{Beijing Institute of Technology}{China}
\affiliation{Institute of Automation}{Chinese Academy of Sciences}{China}
\affiliation{}{University of California, Berkeley}{USA}
\affiliation{}{Tencent AI Lab, Seattle}{USA}

\email{liangwei@bit.edu.cn, lichenxing007@gmail.com, dongyu@ieee.org}

\usepackage{comment}

\begin{document}

\maketitle
\begingroup
\renewcommand\thefootnote{*}
\footnotetext{Equal contribution}
\renewcommand\thefootnote{$\dagger$}
\footnotetext{Corresponding author}
\renewcommand\thefootnote{}
\footnote{Project: https://sites.google.com/view/diffusion-audio-captioning}
\endgroup

\setstretch{0.91}
\begin{abstract}
We introduce \acf{dac}, a non-autoregressive diffusion model tailored for diverse and efficient audio captioning. Although existing captioning models relying on language backbones have achieved remarkable success in various captioning tasks, their insufficient performance in terms of generation speed and diversity impedes progress in audio understanding and multimedia applications. Our diffusion-based framework offers unique advantages stemming from its inherent stochasticity and holistic context modeling in captioning. Through rigorous evaluation, we demonstrate that \ac{dac} not only achieves superior performance levels compared to existing benchmarks in the caption quality, but also significantly outperforms them in terms of generation speed and diversity. 
\end{abstract}

\section{Introduction}

Audio captioning involves detecting sound events and describing acoustic scenes using natural language. The community has witnessed remarkable achievements in audio captioning through \ac{ar} models. Traditional encoder-decoder architectures \cite{narisetty:2022:icassp, xie:2022:icassp, mei2024wavcaps, kim2023prefix, deshmukh2023pengi} use audio encoders to extract audio features and leverage language decoders to generate coherent descriptions. More recently, \ac{llm}-based multimodal models \cite{chu2023qwen, Qwen2-Audio} have emerged, driven by their superior captioning quality and diversity, thanks to a powerful language foundation. However, there are several minor yet non-negligible challenges associated with these models. Encoder-decoder-based models have a lower performance upper bound and can fall into the trap of generating monotonous and repetitive sentences due to their weaker decoders. \ac{llm}-based models are more powerful but require significantly more data and computational resources for training and have slower inference speeds due to the \ac{ar} process. They may also suffer from hallucination problems \cite{bang2023multitask}.

In left-to-right generation tasks, such as text-to-audio or video-to-audio, diffusion models have emerged as a promising approach \cite{ghosal2023text, liu2024audioldm}, offering high-quality outputs and diversity. Furthermore, diffusion-based frameworks, due to the inherent advantages of \ac{nar} models, excel at capturing the target-source dependency \cite{ren2020study, xu2024prompt, ren2025sta}, resulting in a stronger connection between the input source media and the generated output. Although the \ac{nar} structure is typically thought unsuitable for generating inner-coherent contents like text, it facilitates faster generation speed due to parallel decoding and increased diversity due to stochastic noise sampling.

A recent concurrent work DAC-RLD \cite{zhu2025diffusion} leverages diffusion-based structure plus an AR decoder from Bart \cite{lewis2019bart}. The output decoding still follows the paradigm of the AR language model. As the fusion of AR and NAR, the generation is also affected and controlled by the AR decoder. We aim to extend diffusion-based NAR frameworks to audio captioning. In recent works, researchers have designed the pipeline for generating discrete text sequences based on various types of inputs as conditions. Specifically, Denoiser \cite{ye2023dinoiser} facilitates diffusion models for discrete sequence generation by manipulating noises; LaDiC \cite{wang2024ladic} revisits the advantages of diffusion models and highlights their competitiveness in image-to-text generation compared to \ac{ar} models; Prefix-diffusion \cite{liu2023prefix} proposes a lightweight diffusion model for diverse image captioning. A branch of works has demonstrated several key advantages of diffusion framework in text generation: 1) holistic context modeling, where models can capture more overall content rather than inner word relations; 2) parallel decoding, where tokens in the sequence are decoded in parallel and 3) diverse generation, which is brought from the diffusion models. 

We propose Diffusion-based Audio Captioning (DAC), a diffusion-based model for efficient audio captioning. Building on research in image captioning and diffusion-based generation \cite{li2022diffusion, liu2023prefix}, DAC is a pure NAR model and operates in the continuous text latent space. Text descriptions are tokenized, embedded, and mapped into continuous word vectors. Audio is converted to a Mel Spectrogram, encoded by a pre-trained audio encoder, and projected into feature space. The forward process adds noise to the text latent, while the backward process uses the diffusion model to predict noise at each step, conditioned on projected audio features via cross-attention. After further post transition, the text latent is decoded into discrete tokens through a mapping model in a parallel way.

\begin{figure*}[htbp]
\centering
  \centerline{\includegraphics[width=0.9\linewidth]{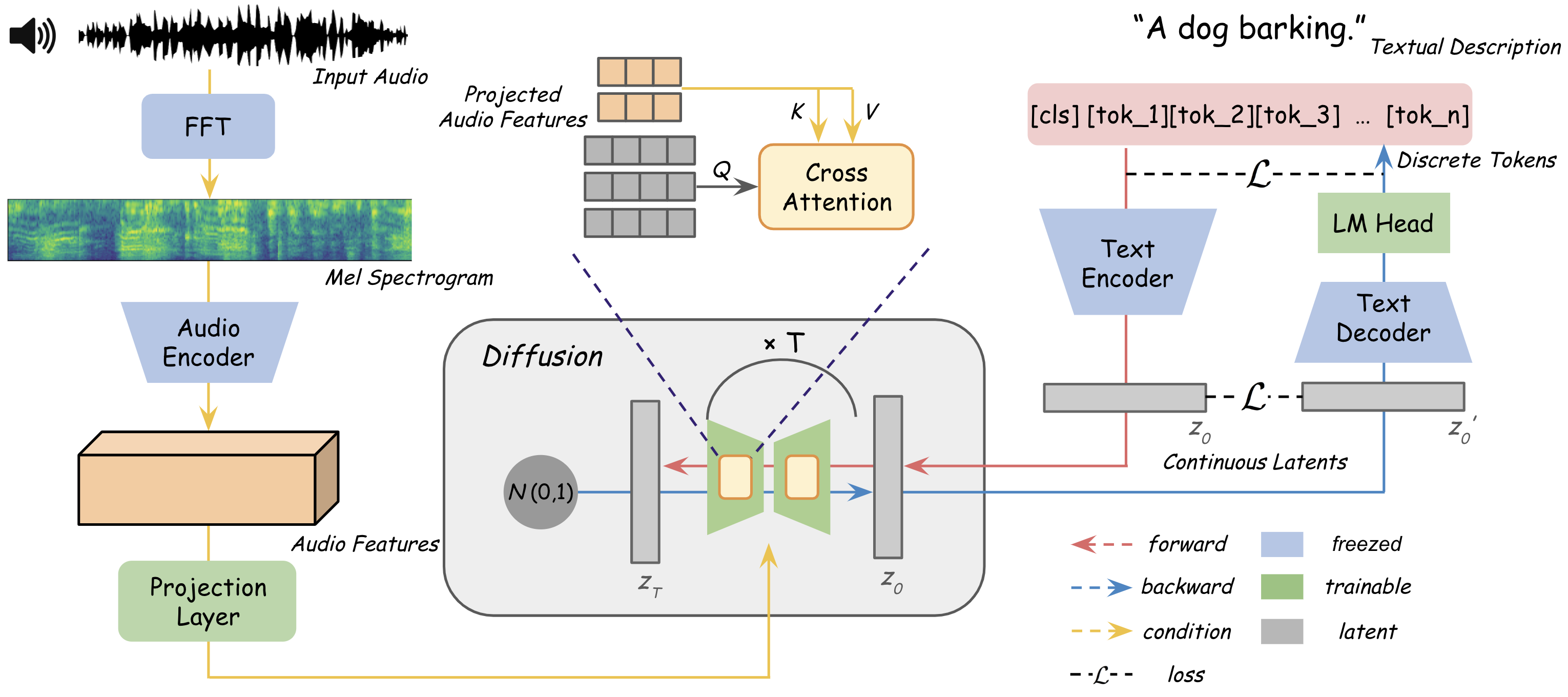}}
  \caption{
  \textbf{An overview of the proposed \ac{dac} framework.} The audio is converted to the audio feature and then as the generation condition through cross attention. The diffusion backbone works on the continuous text latent space and then discrete into tokens.   
  }
  \label{fig:framework}
  \vspace{-10pt}
\end{figure*}

Through evaluation, we demonstrate that \ac{dac} is not only competitive in terms of generation quality compared to SOTA baselines but also surpasses several AR methods in terms of generation diversity and speed. We also provide a further discussion of the commonly-used metrics in captioning tasks, revealing that \ac{dac}'s capabilities extend beyond these metrics. We incorporate extra semantic metrics such as CLAP \cite{wu2023large}, BERT \cite{devlin2018bert}, and GPT4-eval to highlight \ac{dac}'s advantages. 

\section{Diffusion-based Audio Captioning}

\subsection{Diffusion Preliminaries}
\ac{dac} is based on the \ac{ddpm} \cite{ho2020denoising}, where a forward process that repeatedly adds noise sampled from $ N(0, I)$ to the input data:

\begin{equation}
    x_t = \sqrt{1 - \beta_t} x_{t-1} + \sqrt{\beta_t} z_t,
\end{equation}
where $x_0 \sim q$ is the probability distribution to be learned, and then use an estimated function to estimate the undone process of each step in the forward diffusion process:

\begin{equation}
p_{\theta}(x_{T}) = N(x_{T}|0,I),
\end{equation}
\begin{equation}
p_{\theta}(x_{t-1}|x_{t}) = N(x_{t-1}|\mu_{\theta}(x_{t},t),\Sigma_{\theta}(x_{t},t)),
\end{equation}
where the estimated function is parametrized by $\theta$, taking in two arguments $x_{t},t$ and outputting a vector 
$\mu _{\theta }(x_{t},t)$ and a matrix $\Sigma_{\theta(x_{t},t)}$. Maximum likelihood estimation with variational inference is commonly used to optimize the entire process. Traditional diffusion models use U-Net as the underlying architecture to predict noise, while modern models also employ transformer-based structures like DiT \cite{peebles2023scalable} and UViT \cite{bao2023all}. The introduction of transformers allows for cross-attention as an effective way of incorporating conditions during generation.

\subsection{Discrete Text Diffusion Model}




Diffusion models mainly work on the continuous latent space, while textual descriptions are discrete tokens. Works like D3PM \cite{austin2021structured} work on one-hot discrete vector input, and use a categorical distribution and a transition matrix to model the transition probability. We adopt another branch of works like Diffusion-LM \cite{li2022diffusion} and SSD-LM \cite{han2022ssd}. The proposed framework DAC works on the continuous diffusion space. DAC has an extra embedding step in which discrete textual tokens $d = {d_i}$ are embedded to continuous embedding $E(d)$:

\begin{equation}
    q_\phi(x_0|d) = N(E(d), \sigma_0 I), 
\end{equation}
and a final rounding step de-embed the continuous latent variable to discrete tokens:

\begin{equation}
    p_\theta(d|x_0) = \prod_{i=1}^n p_\theta(d_i|x_0).
\end{equation}
DAC uses BERT\footnote{https://huggingface.co/google-bert/bert-base-uncased} \cite{devlin2018bert} as the text encoder. The text decoder consists of two components, as shown in Fig. \ref{fig:framework}: an embedding transition module that maps latent representations to the textual space, and a trainable language model (LM) head that converts embeddings into fluent text. For the transition module, we adopt the design from LaDiC \cite{wang2024ladic}, using BERT to construct the module, with weights initialized from the top layers. The LM head primarily consists of BERT PreTraining Heads, along with additional linear layers, which serve as the generator for discrete tokens. The diffusion module is implemented in two versions: a UViT-based and a DiT-based framework. 

Similar to image captioning works \cite{ren2020study, wang2024ladic,ye2023dinoiser,liu2023prefix}, \ac{dac} primarily incorporates three types of losses: a Mean Squared Error loss that measures the discrepancy between the original latent $x_0$ and the denoised latent $x'_0$, a Cross Entropy loss that evaluates the alignment between the ground truth caption and the generated caption, and an auxiliary valid token loss that aids in constructing the output sequence. We have customized the loss function $\mathcal{L}$ to effectively balance the fitting of the denoised latent $x'_0$ and the generation of the final textual description. 

\begin{table*}[ht]
\centering
\rowcolors{2}{LightCyan1}{}
\caption{Evaluation of \ac{dac} versus SOTA baseline methods on the captioning QUALITY (higher is better)}
\resizebox{0.9\linewidth}{!}{

\begin{tabular}{ccccccccccc} \toprule
 &\multicolumn{10}{c}{\textit{\textbf{Quality}}} \\ 
 \cline{2-11} 
 Baseline & BLEU\_1 & BLEU\_4 &METEOR&Rouge&CIDEr& SPICE & SPIDEr & CLAP & BERT & GPT4-eval\\\midrule
 ACT \cite{Mei2021act} & 0.647 & 0.252 & 0.222 & 0.468 &   0.679 & 0.160 & 0.420 & 0.501 & 0.511 & 6.88            \\ 
 HTSAT-BART \cite{mei2024wavcaps} & 0.675 &  0.272 &  0.237  & 0.483  &  0.711 & 0.177 & 0.444 & 0.522 & \textbf{0.537} & 7.21 \\
   Prefix \cite{kim2023prefix} & \textbf{0.713} & \textbf{0.309}  & 0.240  & \textbf{0.503} &  0.733 & 0.177 & 0.455 & \textbf{0.534} & 0.524  & 7.23 \\ 
 Pengi \cite{deshmukh2023pengi} & 0.691 & 0.253 &  0.232 &   0.482  & 0.752 & 0.182 & 0.467 & 0.473 & 0.524  & 5.21 \\
    DAC-RLD \cite{zhu2025diffusion} & 0.671 & 0.279 & \textbf{0.255} &  0.497 &  \textbf{0.755} & \textbf{0.187} & \textbf{0.471} & 0.531 & 0.521 & 7.12 \\
  Audio-Flamingo \cite{kong2024audio} & 0.449 & 0.079 & 0.191  & 0.344  & 0.266 &  0.124 & 0.195 & 0.461 & 0.484  & 5.76 \\
   Qwen-audio \cite{chu2023qwen} & 0.653 &  0.211 & 0.236 & 0.464  & 0.581 & 0.168 & 0.374  & 0.516  & 0.508 & \textbf{7.60} \\
    Qwen2-audio \cite{Qwen2-Audio} & 0.647 &  0.208 & 0.212 & 0.467  & 0.564 & 0.171 & 0.369  & 0.521  & 0.538 & 7.36 \\
   \midrule
  DAC (HTSAT, UViT)  & 0.634 & 0.220  & 0.215  & 0.457 &  0.627  & 0.145  & 0.386 & 0.527 & 0.511 & 6.93 \\
  DAC (HTSAT, DiT)  & 0.638 & 0.235  & 0.223  & 0.449 &  0.631  & 0.144  & 0.392 & 0.522 & 0.506 & 6.72 \\
  \cline{1-11}
  DAC (BEATs, UViT) & 0.672  &  0.221 & 0.226 & 0.460  & 0.611  &  0.154 & 0.382 & \textbf{0.553} &  0.508  & 7.11   \\ 
   DAC (BEATs, UViT, pt) & \textbf{0.711} &  \textbf{0.295} & \textbf{0.251} & \textbf{0.492}  & \textbf{0.655} & \textbf{0.172} & \textbf{0.414} & 0.548 & \textbf{0.536} & \textbf{7.37}\\
  DAC (BEATs, DiT) & 0.674  &  0.233 & 0.226 & 0.455  & 0.635  &  0.148 & 0.404 & 0.546 &  0.511  & 6.92   \\ 
   DAC (BEATs, DiT, pt) & \textbf{0.713} &  \textbf{0.298} & \textbf{0.253} & \textbf{0.488}  & \textbf{0.674} & \textbf{0.178} & \textbf{0.421} & \textbf{0.549} & \textbf{0.529} & \textbf{7.21}\\\bottomrule 
\end{tabular}
}
\label{tab:results}
\end{table*}

\subsection{Audio Conditioning}
\ac{dac} encodes audio information into the denoising process through cross-attention, where textual vectors serve as queries to retrieve hidden features from the audio latent space. We primarily leverage BEATs \cite{chen2022beats} and HTSAT \cite{chen2022hts} as our audio feature encoder and use a projection module $\psi(\cdot)$ consisting of linear and layer normalization layers to map the audio features $A$ to the audio latent space. The projected audio features $\psi(A)$ are introduced into the denoising process through cross-attention, which is integrated into each block of the denoising net. Mathematically, the cross-attention in \ac{dac} can be represented as:

\begin{equation}
C_t^c = \text{Softmax}\left(\frac{Q_{x_t} \cdot K_{\psi(A)}}{\sqrt{d}}\right) \cdot V_{\psi(A)}, 
\end{equation}
where $Q_{x_t}$ is a projection of the continuous text embedding $x_t$, $K_{\psi(A)}$ and $V_{\psi(A)}$ are different projections of the projected audio features $\psi(A)$, and $d$ is the feature dimension of $K_{\psi(A)}$. The cross-attention mechanism allows text prompts to guide the generation process. We also use classifier-free guidance to help improve the overall captioning quality \cite{ho2022classifier}. During the inference time, the denoising process is performed both conditionally and unconditionally and then extrapolated according to a given weight $w$ called guidance scale: 
\begin{equation}
\tilde{\epsilon}_\theta = w \cdot \epsilon_\theta(x_t, t, \psi(A)) + (1 - w) \cdot \epsilon_\theta(x_t, t, \varnothing),
\end{equation}
where $\varnothing$ demotes the null audio feature.

\section{Experiments}

\subsection{Datasets}
We train our model and compare it with other baselines on the AudioCaps \cite{kim2019audiocaps} dataset, which contains about 46k training instances, 447 validation instances and 957 evaluating instances. In the pre-training experiments, we also leverage Wavcaps \cite{mei2024wavcaps} and Audioset \cite{gemmeke2017audio} as extra training sets, which contain over 2 million training instances.

\subsection{Baselines}
We compare \ac{dac} with several open-source SOTA \ac{ar} audio captioning models. Specifically, we compare encoder-decoder-based models like ACT \cite{Mei2021act}, HTSAT-BART \cite{mei2024wavcaps}, Pengi \cite{deshmukh2023pengi}, and Prefix \cite{kim2023prefix}. At the billion-level parameter size, we compare \ac{llm}-based models like Audio-Flamingo \cite{kong2024audio}, Qwen-audio \cite{chu2023qwen} and Qwen2-audio \cite{Qwen2-Audio}. We also compare diffusion-based concurrent work DAC-RLD \cite{zhu2025diffusion}.

For \ac{dac} models, we report two \ac{dac} models with different audio feature extractors: HTSAT (HTSAT-Audioset), and BEATs (BEATs-ft). We also evaluate a \ac{dac} model that is firstly pre-trained on the Audioset and Wavcaps datasets, and then fine-tuned on AudioCaps.

\subsection{Metrics}
We evaluate several key aspects of the captioning capability of \ac{dac} compared with other baselines. Traditional metrics focus on token-level matching, including BLEU, ROGUEl, METEOR,
CIDEr, SPICE, and SPIDEr. Beyond these common metrics, we also use BERT (bert-base-uncased) \cite{devlin2018bert} and CLAP\footnote{https://huggingface.co/lukewys/laion\_clap/blob/main/630k-best.pt} \cite{wu2023large} to compute the embedding similarity between text-text and text-audio as an extra evaluation of the overall semantic of the generated captions. Following the tendency in \ac{llm} evaluation, we further use GPT-4 (gpt-4-0613) to evaluate the generated caption quality. For the diversity, we mainly use two metrics: lexical-diversity (specifically, a measure of textual lexical diversity, MTLD \cite{mccarthy2010mtld}) and Distinct-N \cite{li2015diversity} (specifically, Distinct-1). MTLD measures the diversity of the textual lexicon, while Distinct-N measures the diversity of a sentence. We use \ac{tps} and \ac{aps} to measure the overall captioning speed of the models.

\subsection{Experimental Settings}
We train \ac{dac} on the AudioCaps for 80 epochs with a batch size of 128 and a learning rate of 1e-4, including 200 warm-up steps. The base model is built on 2D UViT with 6 layers for each encoder and decoder, and the DiT model is based on a 12-layer Transformer with 768 hidden sizes and 12 attention heads. We use Adam as the optimizer and train on 8×NVIDIA A100 GPUs. During inference, we set the skip step to 60 and the guidance scale to 2.5, optimizing parameters via hyperparameter search. For the pre-trained version,  We pre-train \ac{dac} on Audioset and Wavcaps for 60 epochs, and then finetune on AudioCaps for 20 epochs. Inference speed is tested on a single NVIDIA A100-SXM4-40GB. We use the largest possible batch size while loading checkpoints in fp32. Each experiment runs 10 times, and we report the average results.

\section{Results and Analysis}
\subsection{Generation Quality}
As shown in \cref{tab:results}, we present the evaluation results of different captioning quality metrics in AudioCaps\footnote{Baseline scores are taken from original papers or implemented and benchmarked using the official public repositories.}. Comparing \ac{dac} with the other baselines, we demonstrate that although \ac{dac} is not the highest in the rankings among the baselines, it is competitive with the SOTA baselines in most metrics. The standard \ac{dac} model outperforms baselines like ACT and Qwen2-audio and surpasses Pengi and HTSAT-BART in some of the quality metrics. The pre-trained version of \ac{dac} (denoted as pt) outperforms most of the baselines on average across the metrics (We demonstrate that no single model excels in every metric). We would also like to emphasize that since these metrics are built on sequences of tokens, \ac{ar} models inherently possess advantages in these metrics compared to \ac{nar} models. In metrics like CLAP and BERT, the \ac{dac} series of models greatly outperform the other baselines at a large scale, indicating that they have better captured the overall audio features and are more adept at generating holistic descriptions of the given audio. 

Within the \ac{dac} groups, we highlight two key findings: 1) the importance of the audio feature encoder for captioning and 2) the significant impact of pre-training on performance. Using HTSAT, a lightweight audio extractor trained for classification and sound event detection, DAC (UViT, HTSAT) performs competitively with most baselines. Switching to BEATs further improves performance. We also observe that pre-training boosts performance, as shown in the Wavcaps paper. Comparing our results with Wavcaps, we find that pre-training results in a larger improvement in \ac{dac} (e.g., BLEU\_1 increases by 4.74\% for HTSAT-BART and 5.80\% for DAC, while SPIDEr increases by 9.23\% for HTSAT-BART and 24.1\% for DAC). This can be attributed to AudioCaps’ relatively small dataset with a limited language corpus. \ac{ar} models, especially those with LLMs, excel at understanding textual relationships, while diffusion-based models require a larger corpus for effective language learning.

\begin{table}[t]
\centering
\rowcolors{2}{LightCyan1}{}
\caption{Evaluation of \ac{dac} versus SOTA baseline methods on the captioning DIVERSITY and EFFICIENCY (higher is better)}
\resizebox{\linewidth}{!}{
\begin{tabular}{ccccccc} \toprule
 & & \multicolumn{2}{c}{\textit{\textbf{Diversity}}} &  & \multicolumn{2}{c}{\textit{\textbf{Efficiency}}} \\ \cline{3-4} \cline{6-7} 
  Baseline & Param & MTLD & Distinct-1 &     &   TPS& APS        \\\midrule
 ACT \cite{Mei2021act} (beam=2) & 384M &  13.41 & 0.086 &           &  73.26 & \textbf{8.97} \\
  HTSAT-BART \cite{mei2024wavcaps}  &  505M & 13.12 & 0.078 &           & 67.92 &  5.78       \\
Pengi \cite{deshmukh2023pengi}  &  550M & 13.05 & 0.059 &   & \textbf{89.48}   & 5.96        \\
DAC-RLD \cite{zhu2025diffusion} & 372M & 14.08 & 0.089 &  &  63.42 & 6.97 \\
Audio-Flamingo \cite{kong2024audio} & 2.2B & 14.39 & 0.061 &           & 12.99 & 1.26 \\
 Qwen-audio \cite{chu2023qwen} &  7.7B & \textbf{15.08} & \textbf{0.109} &  &          23.12 & 2.46 \\
 Qwen2-audio \cite{Qwen2-Audio} &  7.7B & 14.67 & 0.089 &  & 25.28 & 2.43 \\
\midrule
 DAC (HTSAT, UViT)  & 573M & 14.14 & \textbf{0.104} &           &  \textbf{49.28} & 6.48 \\
 DAC (HTSAT, DiT)  & 597M & 14.12 & 0.098 &           &  \textbf{67.32} & 8.29 \\
\cline{1-7}
 DAC (BEATs, UViT) & 627M & \textbf{14.53} & \textbf{0.104} &           & 46.60 & 5.88 \\ 
 DAC (BEATs, UViT, pt) & 627M & 14.45 & 0.100 &           & 47.52 & \textbf{6.64} \\
 DAC (BEATs, DiT) & 651M & 14.50 & 0.101 &           & 60.28 & 8.03 \\ 
 DAC (BEATs, DiT, pt) & 651M & \textbf{14.53} & \textbf{0.102} &           & 60.23 & \textbf{8.12} \\
 \bottomrule 
\end{tabular}}
\label{tab:results-2}
\end{table}

\subsection{Diversity and Efficiency}

We present the diversity and efficiency evaluation results in \cref{tab:results-2}, where we also report the model parameter size. We demonstrate that \ac{dac} models achieve significant diversity at both the lexical and sentence levels, while maintaining a relatively small model size and fast generation speed. In terms of diversity, only the Qwen-audio series surpass \ac{dac}, but with a much larger parameter size and a much slower inference speed. A stronger LLM backbone can indeed help increase generation diversity and quality, but this comes at the cost of greater computational resources and slower generation speed. Regarding generation efficiency, the fastest model, ACT, is almost eight times faster than the slowest Audio Flamingo with the same computational resources. Our \ac{dac} models are also significantly faster than \ac{ar} models while maintaining competitive performance. Compared to DAC-RLD, our model is competitive in generation speed but outperforms it in diversity. We attribute this to the DAC-RLD design, where the subsequent AR decoder from the BART module may mask the potential of diffusion in enhancing diversity.

Beyond the quantitative results, we observe two advantages that may not be evident from the table. Firstly, \ac{dac} is not sensitive to caption length due to their holistic denoising and decoding process on embeddings, as long as the length does not exceed the model's token limit. This benefits both the generation speed and overall generation quality. For \ac{dac}, the generation token length is limited to 40, while the average caption length for AudioCaps is around 10. In contrast, \ac{ar} models experience a linear increase in generation time with sequence length. Longer sequences may also lead to distraction challenges in \ac{llm}s. Secondly, a smaller parameter size facilitates easier deployment on devices or allows for larger batch sizes. 

\begin{figure}[t]
\centering
  \centerline{\includegraphics[width=0.9\linewidth]{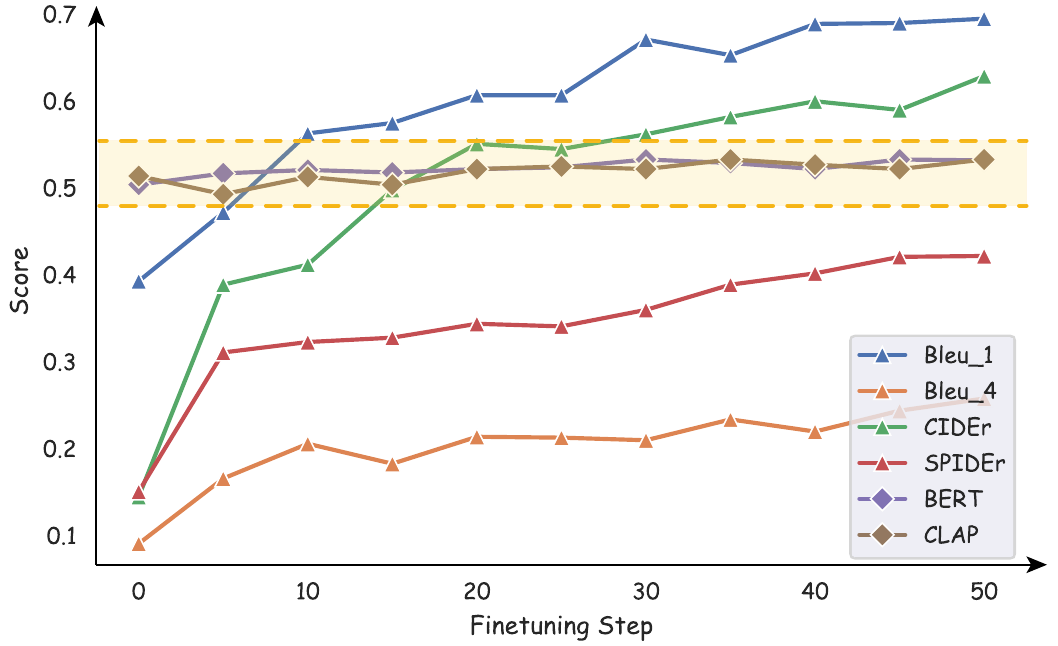}}
  \caption{
  \textbf{Change of the metric scores on the AudioCaps dataset during fine-funing following pre-training.} We demonstrate that achieving good convergence during Audioset pre-training does not necessarily translate to a strong initial performance on the AudioCaps test set, especially for metrics like BLEU and SPIDEr. }
  \label{fig:metrics}
\end{figure}

\subsection{Further Discussion about the Metrics}

An intriguing observation during the \ac{dac} fine-tuning process on AudioCaps is that achieving good convergence during the pre-training phase, \textit{e.g.}, pre-training on Audioset, does not guarantee a strong initial performance on the AudioCaps test set for certain metrics, as illustrated in \cref{fig:metrics}. Although pre-training on a significantly larger dataset yields much better results from the loss perspective (around 5 for Audioset \textit{vs} around 7 for AudioCaps), metrics such as BLEU, CIDEr, and SPIDEr are considerably lower than the baseline level if we directly evaluate the checkpoint on the AudioCaps dataset. In contrast, metrics like CLAP and BERT are maintained consistently.

We thus consider metrics such as CLAP and BERT to be more equitable and convincing scores for evaluating captioning ability in our experiment. Traditional metrics like BLEU and CIDEr are based on token-level matching, implying that a better imitation of the ground truth strings results in a higher score. However, our findings suggest that a good caption does not necessarily have to replicate the ground truth verbatim. As demonstrated in \cref{fig:metrics}, there are instances where the generated captions are semantically similar to both the audio and the ground truth text, yet they receive low scores due to differences in caption style or grammar. Fine-tuning within the target distribution also does not necessarily enhance overall semantic similarity. On the other hand, metrics like CLAP and BERT measure the overall semantic similarity. On these two metrics, our model outperforms the other baselines.

\section{Conclusion}

We present DAC, a diffusion-based NAR model for generating diverse and high-quality audio captioning. Leveraging the \ac{nar} architecture, \ac{dac} excels in providing a superior mix of diversity, efficiency, and quality in generating textual descriptions of audio pieces, outperforming several existing \ac{ar} models. Our model surpasses current benchmarks, especially in terms of generation diversity and processing speed, while maintaining a lightweight design. We hope that our findings will inspire further exploration of a diffusion-based comprehensive framework for multimodal content generation.

\newpage
\setstretch{0.8}

\bibliographystyle{IEEEtran}
\bibliography{mybib}

\end{document}